\documentclass[conference]{IEEEtran}
\IEEEoverridecommandlockouts

\usepackage{amsthm,amsmath}
\usepackage[utf8]{inputenc}
\usepackage{import}
\usepackage{graphicx}
\usepackage{rotating}
\usepackage[graphicx]{realboxes}
\graphicspath{ {images/} }
\usepackage{array}
\usepackage[citestyle=numeric-comp,bibstyle=ieee,%
backend=biber,sortcites]{biblatex}
\usepackage{orcidlink}

\begin{document}




\title{Interpretable Deep Learning for Forecasting Online Advertising Costs: Insights from the Competitive Bidding Landscape
\thanks{This work was funded by Fundação para a Ciência e a Tecnologia (UIDB/00124/2020, UIDP/00124/2020 and Social Sciences DataLab - PINFRA/22209/2016), POR Lisboa and POR Norte (Social Sciences DataLab, PINFRA/22209/2016).}}

\author{\IEEEauthorblockN{Fynn Oldenburg}
\IEEEauthorblockA{\textit{Department of Operations \& Technology} \\
\textit{Kühne Logistics University}\\
Hamburg, Germany \\
fynn.oldenburg@klu.org}
\and
\IEEEauthorblockN{Qiwei Han\orcidlink{0000-0002-6044-4530}}
\IEEEauthorblockA{\textit{Nova School of Business and Economics} \\
\textit{Universidade NOVA de Lisboa}\\
Carcavelos, Portugal  \\
qiwei.han@novasbe.pt}
\and
\IEEEauthorblockN{Maximilian Kaiser\orcidlink{0009-0007-4329-161X}}
\IEEEauthorblockA{\textit{
        Chair of Marketing \& Media } \\
\textit{University of Hamburg}\\
Hamburg, Germany \\
maximilian.kaiser@uni-hamburg.de}
}

\maketitle
\begin{abstract}

As advertisers increasingly shift their budgets toward digital advertising, accurately forecasting advertising costs becomes essential for optimizing marketing campaign returns. This paper presents a comprehensive study that employs various time-series forecasting methods to predict daily average CPC in the online advertising market. We evaluate the performance of statistical models, machine learning techniques, and deep learning approaches, including the Temporal Fusion Transformer (TFT). Our findings reveal that incorporating multivariate models, enriched with covariates derived from competitors' CPC patterns through time-series clustering, significantly improves forecasting accuracy. We interpret the results by analyzing feature importance and temporal attention, demonstrating how the models leverage both the advertiser's data and insights from the competitive landscape. Additionally, our method proves robust during major market shifts, such as the COVID-19 pandemic, consistently outperforming models that rely solely on individual advertisers' data. This study introduces a scalable technique for selecting relevant covariates from a broad pool of advertisers, offering more accurate long-term forecasts and strategic insights into budget allocation and competitive dynamics in digital advertising.
\end{abstract}







\section{Introduction}
The global advertising market has evolved beyond recognition over the past decades. According to Dentsu’s latest Global Ad Spend Forecast report, the total ad spend worldwide is expected to reach \$752.8 billion in 2024, with digital advertising accounting for 58.8\% \cite{dentsu2023}. Despite this growth, many small to medium-sized advertisers face challenges due to limited technical expertise and marketing resources. These advertisers often collaborate with digital advertising platforms like Google and Facebook, which use auction mechanisms that allow them to bid on keywords for paid search advertising. Accurately forecasting advertising costs, particularly cost-per-click (CPC), is essential for these advertisers to optimize their budget plans and estimate the returns on their pay-per-click (PPC) campaigns \cite{evans2008economics,zia2019search}.

Overall, the online advertising cost is dynamically determined by multiple factors, such as macroeconomic conditions \cite{greenwood2021you}, industry \cite{bagwell2007economic}, the quality of the ads \cite{shen2023price}, and the keywords \cite{rutz2011modeling}, among others. For example, according to Worldstream, the average CPC in Google Ads can vary dramatically across sectors, with the legal industry and consumer services seeing CPCs above \$6, while e-commerce averages around \$1 \cite{wordstream2022}. In particular, advertising costs highly depend on the competitive landscape of the online advertising market, i.e., the CPC depends on the competing advertisers’ bidding prices at the time of the auction \cite{evans2008economics}. However, advertisers that only have access to their own budget plans may lack a holistic view of the bidding strategies of other advertisers. For example, Google Ads provides \href{https://ads.google.com/home/tools/keyword-planner/}{\textit{Keyword Planner}}  for advertisers to evaluate their budget plans to forecast the future performance of marketing campaigns. However, such an estimated CPC would only reflect the generic weekly average maximum CPC based on different bidding strategies and would need to be adjusted to generate actual CPC. Additionally, the tool only displays the forecasted CPC, limiting advertisers from gaining insights into the cost structure of the online advertising market.

To address this gap, our study aims to forecast CPC for advertisers across different industries over multiple time horizons, using a large-scale paid search dataset that includes many advertisers engaged in bidding competitions. We emphasize the importance of understanding the competitive landscape by applying advanced time-series clustering techniques to identify relevant covariates from competing advertisers. We employ a range of time-series forecasting methods, including statistical approaches (SARIMA), machine learning techniques (XGBoost), and deep learning models (LSTM and Transformer). For models that can handle covariates, we introduce a comprehensive set of features derived from the competitive landscape, including data on both the advertisers and their competitors. Our results demonstrate that multivariate forecasting enriched with competitive features through distance-based clustering, consistently outperforms univariate forecasting.


Moreover, we show that the Temporal Fusion Transformer (TFT), which leverages covariates obtained from distance-based clustering, achieves superior performance compared to other methods. We interpret the feature importance outputs of the TFT, illustrating how the model captures the non-linear effects of budget levels on CPC. Importantly, our study highlights the benefits for advertisers of understanding the competitive landscape in Google Ads auctions and adopting strategies from similar advertisers. Through a policy experiment, we also demonstrate the robustness of our proposed model during periods of significant economic disruption, further showcasing its advantages over models based solely on individual advertiser data. This underscores the added value of our approach in delivering detailed, interpretable, and strategically valuable CPC forecasts, surpassing the capabilities of existing tools like Google's Keyword Planner.

\section{Data}
\subsection{Data Description}

\begin{figure*}[!htb]
\centering
\includegraphics[width=13.2cm]{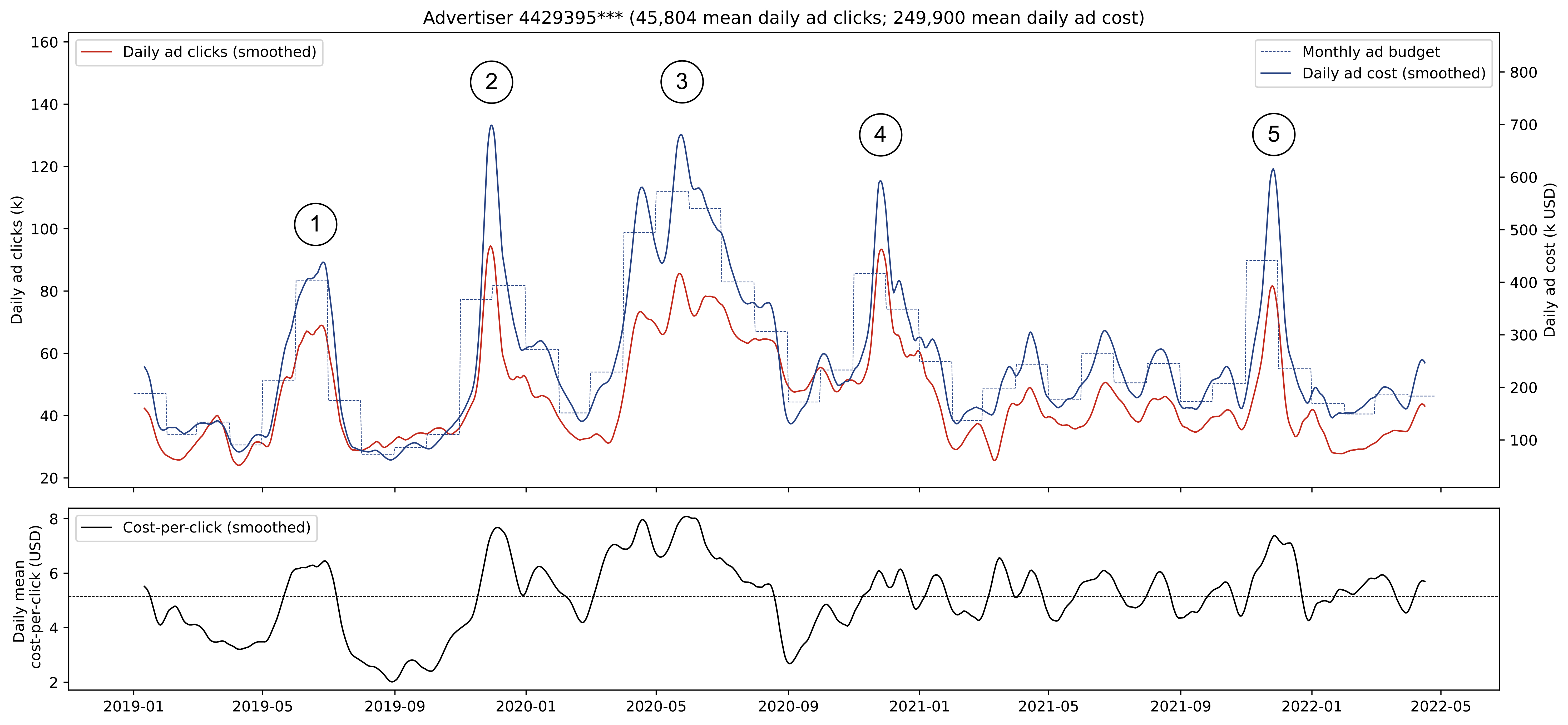}
\caption{An illustrated example of daily ad clicks, cost and CPC from one advertiser} 
\label{fig:feature_logic}
\end{figure*}

The data for this study was provided by European e-commerce research platform \href{https://gripsintelligence.com/}{Grips Intelligence}, which collected ad spend data through global partnerships. The dataset includes daily Google paid search advertising metrics from 2019 to 2022, covering ad costs, clicks, and impressions for over 249,000 anonymized German advertisers. In Google Ads, advertisers set a monthly budget, which Google allocates to individual auctions daily \cite{googleadsense2022}. This allocation varies daily, often reflecting weekly seasonality with fluctuating amplitudes \cite{yuan2013}. We chose to analyze the data on a daily basis to capture the effects of daily bidding fluctuations and competitive actions in the marketplace, providing a more precise forecast of CPC across different time horizons. The dataset also spans the COVID-19 pandemic, which significantly impacted online pricing structures \cite{r2020advertising}. However, since our objective is to forecast CPC—a relative measure of cost per click—the absolute cost changes during the pandemic should not affect our model's robustness. To ensure data quality and reliability, we excluded advertisers with more than one percent of missing daily values. For the remaining data with missing values, we applied linear interpolation, generating data points by connecting the two closest available observations linearly.

The target variable, daily average cost-per-click (CPC), was derived by dividing the actual daily ad cost by the number of ad clicks on a given day. Additionally, we created several temporal descriptive features to support the modeling process, including day of the week, day of the year, month of the year, and a 7-day lag of CPC. These features are critical for capturing weekly seasonality and identifying special days like Christmas, which can significantly influence advertising patterns.

\subsection{Exploratory Data Analysis}

Understanding the interplay between key variables in online advertising is essential for constructing robust predictive models. Figure \ref{fig:feature_logic} illustrates the relationships between these variables. An increased ad budget aims to achieve higher ranks in bidding iterations, thereby generating more attention. Reliable measures of ad attention include how often the ad appears on a user’s screen (daily ad impressions) and how often the user clicks on the ad (daily ad clicks). As depicted in Figure 1, the number of clicks generated is highly influenced by the budget input. Across all advertisers in the dataset, clicks and impressions have a Pearson correlation coefficient of 0.71 and 0.69, respectively, with the monthly ad budget.

Moreover, CPC indicates how effectively a campaign converts ad budget into user clicks—higher CPC values suggest less effective marketing. Our example in Figure \ref{fig:feature_logic} shows that the development of CPC cannot be explained solely by the aforementioned variables. Even though budget increases, as noted in Figure 1, the conversion to actual daily ad cost and CPC does not follow proportionally. This suggests that additional factors influencing online advertising effectiveness are embedded within the data from multiple competing entities.

Recent research supports this notion, indicating that various online advertising-related factors impact CPC and actual sales through online channels, with the significance of these variable dependencies fluctuating over time \cite{yang2022}. This highlights the complexity of the online advertising ecosystem and the need for sophisticated models that can account for these dynamic interdependencies.

\section{Methodology}
In this study, we test various model configurations combining predictive models, feature compositions, and clustering methods to identify the best-performing setup for predicting the future CPC of online advertisers. We selected four model types representing different categories of time series forecasting approaches: SARIMA as a statistical model, Extreme Gradient Boosting (XGB) from classical machine learning, Long Short-Term Memory neural networks (LSTM) from deep learning, and the Temporal Fusion Transformer (TFT), representing novel transformer-based architectures.

\begin{figure*}[!htb]
\centering
\includegraphics[scale=0.45]{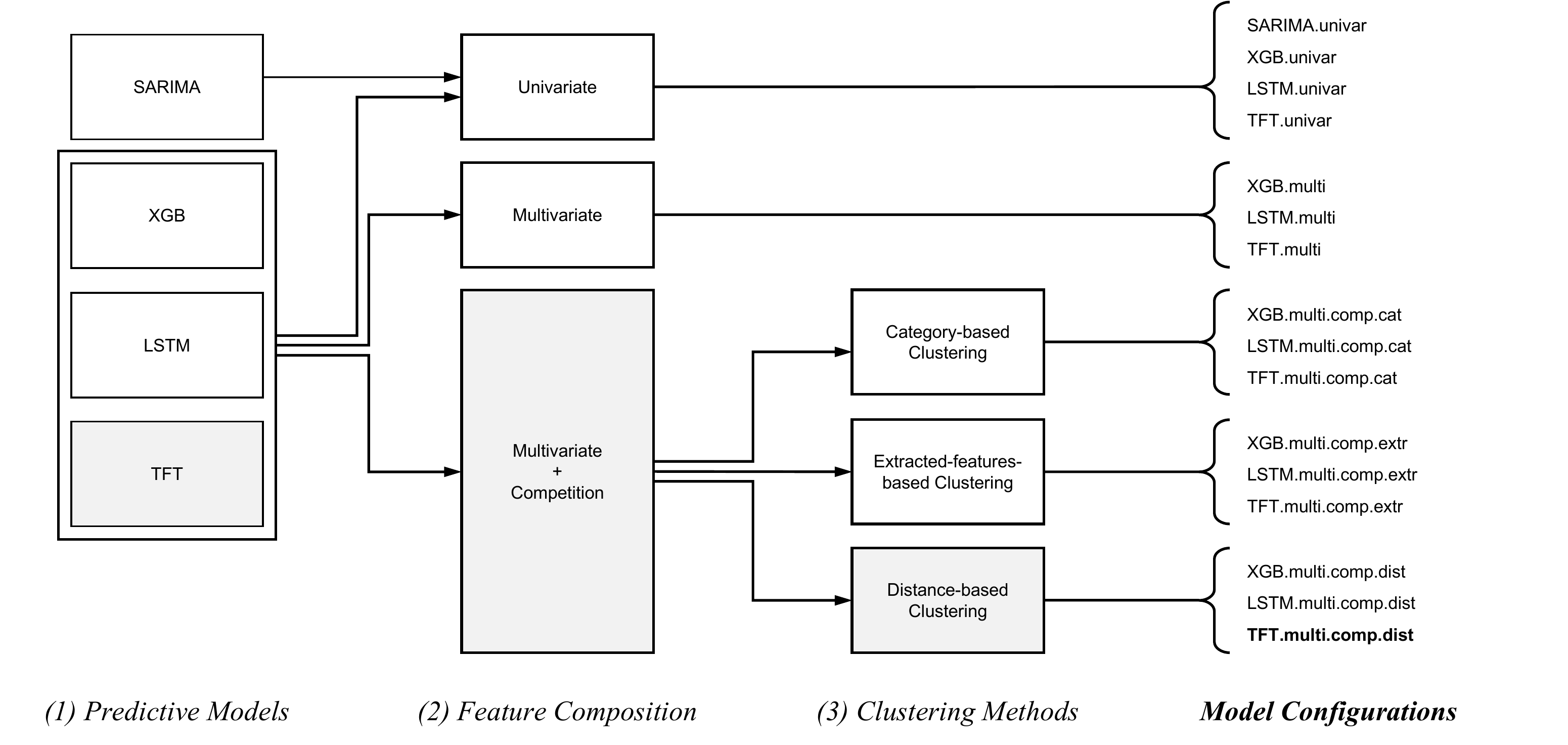} 
    \caption{Study setup showing the resulting configurations of predictive models, feature compositions, and clustering methods. SARIMA is used as a benchmark model and is only tested in a univariate configuration. The best-performing configuration is the Temporal Fusion Transformer based on multivariate features, including competitors' data identified through distance-based clustering (highlighted in grey).}
 \label{fig:study_setup}
\end{figure*}

Except for SARIMA, which is used as a univariate benchmark model, each model type is tested with three different compositions of input features (section \ref{section3.b}): \textit{univariate}, \textit{multivariate} and \textit{multivariate + competition}, where competitions among advertisers are determined through time series clustering (section \ref{section3.c}). Figure \ref{fig:study_setup} illustrates the total of 16 model configurations. Each configuration is run iteratively to produce forecasts for every advertiser. We evaluate performance based on three forecasting horizons: 14 days, 30 days, and 60 days. This allows us to cover short-term forecasting for reacting to current market situations and long-term forecasting for strategic budget planning. Models are compared using aggregated average mean absolute error (MAE) and symmetric mean absolute percentage error (SMAPE) across all advertisers.

\subsection{Predictive Models}

\subsubsection{Temporal Fusion Transformer}
The Temporal Fusion Transformer (TFT) is an attention-based deep neural network architecture introduced by \cite{lim2021temporal}. In addition to accurate predictions across multiple forecasting horizons, TFT provides detailed interpretability of feature importance and temporal dynamics. The TFT consists of multiple processing layers: i) Variable Selection Networks (VSN): Processes all known, unknown, and static inputs, filtering out non-contributing features. ii) LSTM Encoder-Decoder Layer: Enhances locality by processing known and unknown time series with LSTM cells, followed by a decoder fed with known series. iii) Gated Residual Network (GRN): Adds non-linear complexity adjustments for each input. iv) Temporal Self-Attention Layer: Captures long-term trends with multi-head attention for interpreting temporal patterns. TFT has been successfully applied in online advertising for predicting advertising revenues \cite{wurfel2021online}. 

\subsubsection{Long Short-term Memory}
Long Short-Term Memory (LSTM) is a deep learning architecture known for its efficiency in learning temporal patterns with minimal preprocessing \cite{hochreiter1997long, lim2021survey}. LSTM is particularly suited for long time series, making it ideal for our dataset, which includes over a thousand daily observations. Each historical time step is represented in an LSTM cell, which processes the series through gates that retain significant sequence parts while forgetting less important ones. The final hidden state generates a multi-step output vector for forecasting.

\subsubsection{Extreme Gradient Boosting}
Extreme Gradient Boosting (XGBoost) is an implementation of the Gradient Boosting Decision Trees (GBDT) algorithm \cite{chen2016}. XGBoost models consist of multiple shallow decision trees whose combined output yields accurate predictions. For time series forecasting, the dataset is restructured into a machine learning problem, with separate models fitted for each time step in the target sequence. XGBoost’s interpretability stems from feature importance scores aggregated across all decision trees.

\subsubsection{SARIMA}
The Seasonal Autoregressive Integrated Moving Average (SARIMA) is a statistical method for modeling univariate time series. SARIMA extends ARIMA by incorporating seasonal components. It includes three main components: autoregressive (AR), moving average (MA), and differencing to ensure stationarity, alongside additional hyperparameters that describe the nature of seasonality and the number of time steps per seasonal period. Previous studies have shown that statistical models like SARIMA often outperform more complex algorithms for univariate problems, especially on shorter time series, making it a reliable baseline model \cite{makridakis2018, spiliotis2021}.

\subsection{Feature Composition}\label{section3.b}

The cost structures of online advertising are heavily influenced by the competitive behavior of advertisers within Google’s keyword bidding architecture. Higher competition on certain keywords typically results in higher CPC for bidders. We leverage these cost structures to build multivariate forecasting models, incorporating information about competing advertisers.

\begin{figure*}[!t]
\centering
\includegraphics[width=14.2cm]{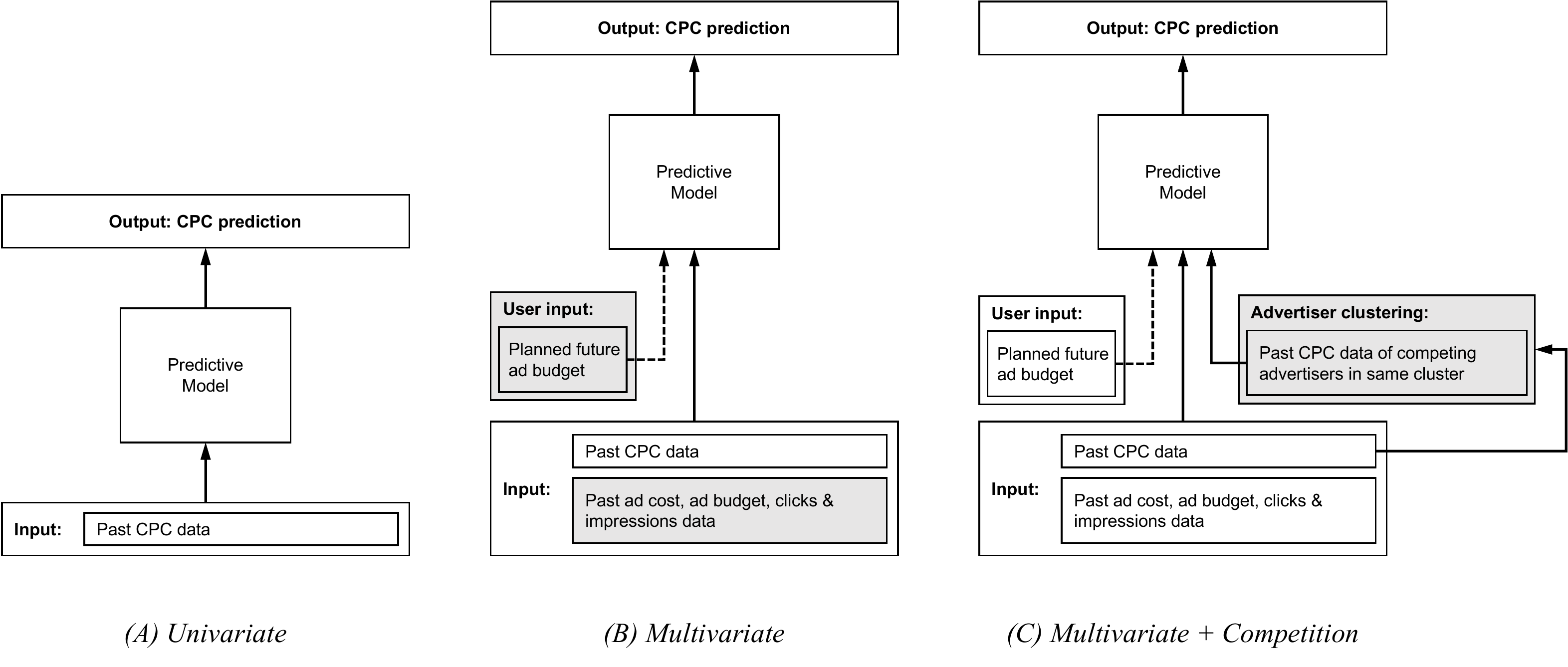}
\caption{Feature composition setup for our three configurations (added component highlighted in grey).}
\label{fig:feature_composition}
\end{figure*}

We experiment with different feature compositions for our models. Figure \ref{fig:feature_composition} shows the three feature sets tested for each model:

\begin{itemize}
    \item Univariate: Includes only historical CPC data for the target advertiser.
    \item Multivariate: Includes historical data on ad cost, budget, clicks, and impressions for the target advertiser, as well as planned ad budget.
    \item Multivariate + Competition: Incorporates features from competing advertisers, grouped using time-series clustering.
\end{itemize}

\subsection{Clustering Methods}\label{section3.c}

\subsubsection{Category-based Clustering}

As the data provides information about the primary category assigend to each advertiser, we select this feature to create baseline clusters. Advertisers from the same industry are natural competitors, likely advertising similar products and services. This category-based clustering serves as a benchmark for comparing our other clustering methods and provides insights into cross-category competition.

\subsubsection{Extracted-features-based Clustering}

There are a variety of ways to extract features for time series clustering. The proposed methods range from generating a large number of features to represent the characteristics of the time series as accurately as possible, to extracting only selected descriptive features in order to avoid irritation from noise and outliers in the data. We follow the approach of \cite{bandara2020} to generate high-level extracted features for time series clustering due to its efficiency. More specifically, this method involves generating 14 descriptive features (Table \ref{extracted_feats}) to represent the characteristics of the time series. The k-Means algorithm, with the number of clusters determined by the elbow method, is then applied to these features.

\begin{table}[!htb]
\centering
\fontsize{6}{8}\selectfont
    {\renewcommand{\arraystretch}{1.15} 
        \begin{tabular}{p{3cm}p{4.5cm}}
        \hline
          \textbf{Extracted feature} & \textbf{Description} \\
          \hline
          mean & Mean \\
          variance & Variance \\
          acf\_1 & First order of autocorrelation \\
          trend & Strength of trend \\
          linearity & Strength of linearity \\
          curvature & Strength of curvature \\
          season & Strength of seasonality \\
          peak & Strength of peaks \\
          trough & Strength of trough \\
          entropy & Spectral entropy \\
          lumpiness & Changing variance in remainder \\
          spikiness & Strength of spikiness \\
          f\_spots & Flat spots using discretization \\
          c\_points & Number of crossing points \\
        \hline
        \end{tabular}
    }
\vspace{0.2cm}
\caption{Set of extracted time series features following \cite{bandara2020}}
\label{extracted_feats}
\end{table}

\subsubsection{Distance-based Clustering}

We use the time series k-Means (TSkmeans) algorithm \cite{huang2016} for distance-based clustering. TSkmeans addresses the problem of matching noisy time series by selecting smooth subspaces and assigning weights to the most valuable timestamps. This algorithm leverages averaging methods introduced by \cite{petitjean2011} to use classical distance measures for serial data. The distance measures used include Euclidean distance and Dynamic Time Warping (DTW).

\begin{figure*}[!t]
\centering
\includegraphics[width=14.2cm]{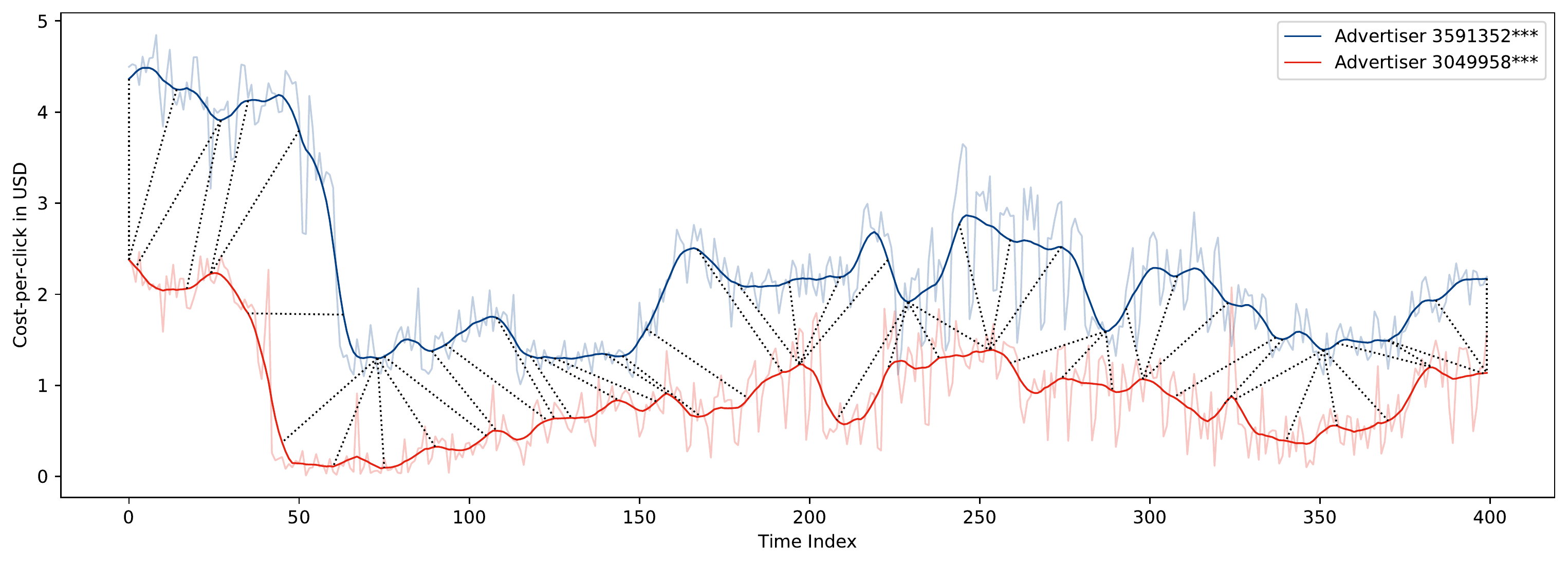}
\caption{Example of matching two smoothed advertiser time series using Dynamic Time Warping as a distance measure.}
\label{fig:dtw}
\end{figure*}

Euclidean distance, though simple and intuitive, assumes aligned time series of the same length, which is often violated in practice. DTW, introduced by \cite{Berndt1994}, solves this by warping time series non-linearly, allowing for differences in time scales and non-uniform sampling rates. DTW calculates the minimum accumulated distance between two time series by moving through a matrix of pairwise distances, ensuring alignment before computing the distance. Figure \ref{fig:dtw} shows how DTW aligns two smoothed advertiser time series with similar patterns occurring at different times and scales.

\section{Results}
\subsection{Model Performance Comparison}
The final performance scores for each model configuration across the three forecasting horizons are presented in Table \ref{table_results}. The overall best-performing model in our evaluation is the Temporal Fusion Transformer (TFT), particularly when including time series from competing advertisers identified through distance-based clustering as covariates. Notably, this configuration outperforms all other models across the three performance measures, especially for the 60-day forecasting horizon. While the univariate SARIMA model performs well for shorter forecasting horizons, its accuracy deteriorates rapidly for longer horizons, where it is outperformed by more complex models that incorporate multivariate features.

For all tested models, the best configuration consistently includes clustered multivariate features from competitors as covariates. The exceptions are the 14-day XGBoost and 14-day LSTM predictions, where category-based and extracted-features-based clustering, respectively, yield the best results. However, in terms of SMAPE, distance-based clustering generally produces the most accurate predictions. This finding contrasts with \cite{bandara2020}, who found that clustering based on extracted features offered better results, suggesting that using actual time series data points and Dynamic Time Warping (DTW) for clustering is more effective in our context. Interestingly, using advertiser categories as clusters was outperformed by both clustering algorithms in nearly every test scenario. This could be due to two factors: First, our dataset includes 21 advertiser categories, compared to the six or seven clusters generated by the clustering algorithms. Models like XGBoost and TFT, which have inherent processes for filtering relevant features, likely benefit from accessing a larger pool of exogenous advertiser time series, increasing the chances of identifying relevant covariates. Second, the algorithm-generated clusters may capture cross-category competitive patterns, which are not as apparent in the native category groupings.

\begin{table*}[!t]
\centering
\fontsize{6}{8}\selectfont
{\renewcommand{\arraystretch}{1.25} 
    \begin{tabular}{p{2.2cm}p{0.02cm}p{0.9cm}p{1.1cm}p{0.02cm}p{0.9cm}p{1.1cm}p{0.02cm}p{0.9cm}p{1.1cm}}
    \hline
    \textbf{Model} & & \textbf{14 days} & & & \textbf{30 days} & & & \textbf{60 days} & \\
    & & MAE $\downarrow$ & SMAPE $\downarrow$ & & MAE $\downarrow$ & SMAPE $\downarrow$ & & MAE $\downarrow$ & SMAPE $\downarrow$ \\
    \cline{1-1}
    \cline{3-4}
    \cline{6-7}
    \cline{9-10}
        SARIMA.univar & & 0.211 & 0.179 & & 0.227 & 0.197 & & 0.306 & 0.264 \\
        \vspace{1pt} \\
        XGB.univar & & 0.259 & 0.225 & & 0.278 & 0.243 & & 0.340 & 0.279 \\
        XGB.multivar & & 0.242 & 0.218 & & 0.271 & 0.231 & & 0.274 & 0.258 \\
        XGB.multivar.comp.cat & & 0.245 & 0.219 & & 0.250 & 0.218 & & 0.278 & 0.423 \\
        XGB.multivar.comp.extr & & 0.248 & 0.228 & & 0.257 & 0.216 & & 0.267 & 0.239 \\
        XGB.multivar.comp.dist & & 0.244 & 0.213 & & 0.246 & 0.227 & & 0.276 & 0.235 \\
        \vspace{1pt} \\
        LSTM.univar & & 0.245 & 0.203 & & 0.265 & 0.217 & & 0.318 & 0.251 \\
        LSTM.multivar & & 0.235 & 0.204 & & 0.250 & 0.215 & & 0.252 & 0.236 \\
        LSTM.multivar.comp.cat & & 0.231 & 0.199 & & 0.229 & 0.202 & & 0.253 & 0.228 \\
        LSTM.multivar.comp.extr & & 0.227 & 0.202 & & 0.235 & 0.199 & & 0.253 & 0.237 \\
        LSTM.multivar.comp.dist & & 0.229 & 0.196 & & 0.231 & 0.210 & & 0.250 & 0.232 \\
        \vspace{1pt} \\
        TFT.univar & & 0.216 & 0.185 & & 0.236 & 0.204 & & 0.307 & 0.241 \\
        TFT.multivar & & 0.203 & 0.180 & & 0.223 & 0.192 & & 0.241 & 0.214 \\
        TFT.multivar.comp.cat & & 0.208 & 0.181 & & 0.215 & 0.188 & & 0.238 & 0.204 \\
        TFT.multivar.comp.extr & & 0.207 & 0.181 & & \textbf{0.211} & 0.186 & & 0.235 & 0.202 \\
        TFT.multivar.comp.dist & & \textbf{0.201} & \textbf{0.177} & & 0.216 & \textbf{0.183} & & \textbf{0.232} & \textbf{0.196} \\
    \hline
    \end{tabular}
}
\vspace{0.2cm}
\caption{Model performance comparison of all model and clustering configurations across three forecasting horizons}
\label{table_results}
\end{table*}

For all models, the univariate configuration shows a sharper decline in performance over longer horizons compared to the multivariate configurations. This could be attributed to the inclusion of budget data as a covariate in the multivariate models. As explained further in Section \ref{section4.3}, CPC development is highly dependent on the advertiser's budget, which is typically set on a monthly basis. Models that have access to budget information and are sufficiently complex to learn the monthly input structure perform better at predicting changes over forecasting horizons longer than a month. Our results support this observation, as univariate models show a significant decrease in prediction performance, similar to univariate statistical models, while multivariate models maintain more stable performance.

Lastly, the TFT model exhibits the smallest performance difference between the three clustering configurations. Transformers are particularly effective when handling large datasets, such as long time series and numerous features. The TFT architecture includes a variable selection network layer that filters the most relevant variables. The model likely assigns the highest weights to a few highly relevant exogenous advertiser time series included in the same cluster, regardless of the clustering approach used.

\subsection{Cluster Assignment}

\begin{figure*}[!t]
\centering
\includegraphics[width=15cm]{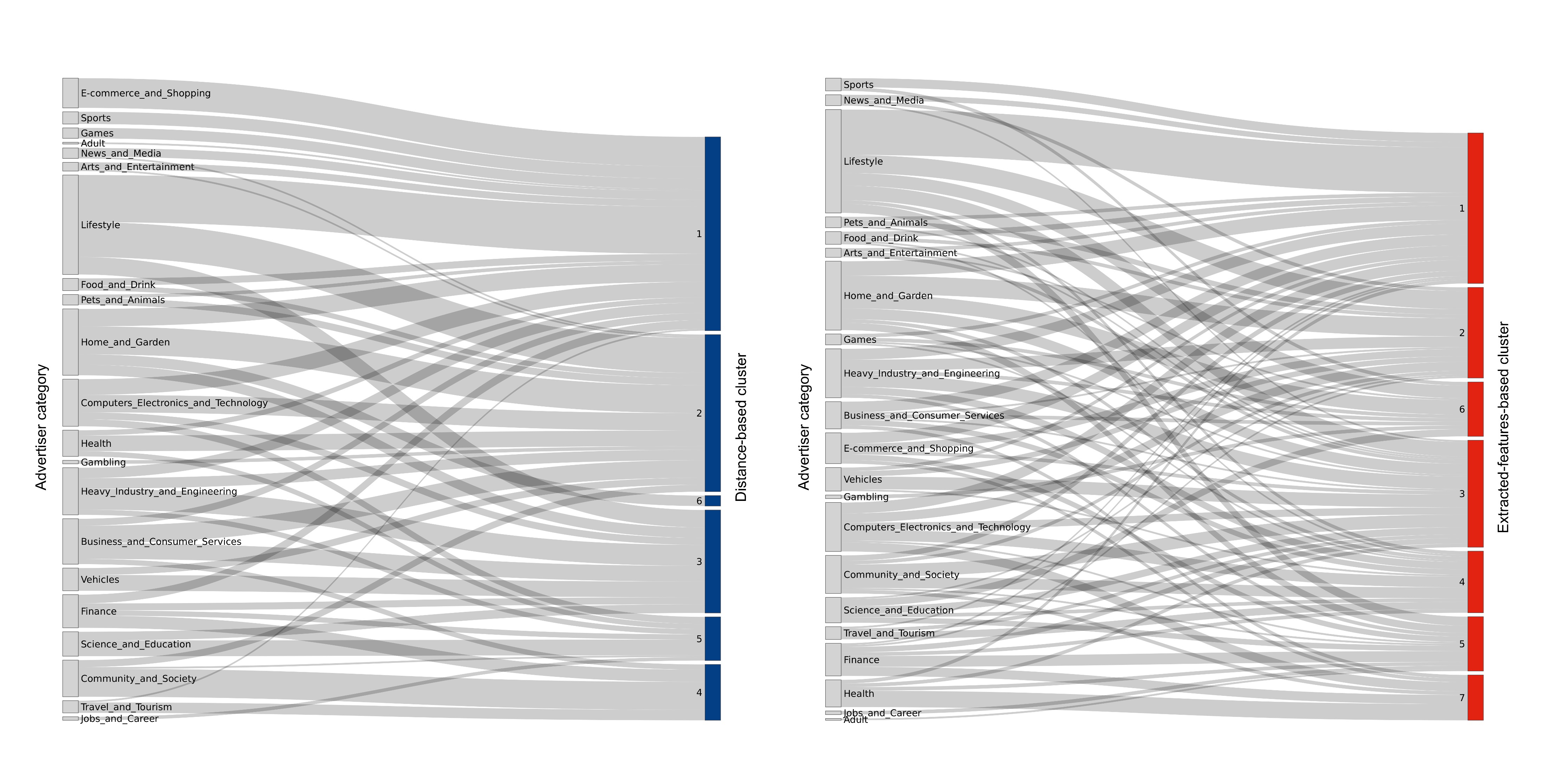}
    \caption{Sankey diagram of cluster assignments using the distance-based (left) and extracted-features-based approach (right) compared to the advertisers' native category.}
 \label{fig:sankei_distance}
\end{figure*}

Our two clustering assignments, generated by applying k-Means to a distance-based representation and an extracted-feature-based representation of advertisers' CPC time series, differ from each other and from the advertiser categories we use as baseline clusters. Following the elbow method, the distance-based approach assigns advertisers to six clusters. There is a strong representation of advertiser categories in the assignment, with larger groups of advertisers from the same category often located in the same clusters. More than half of the advertisers are assigned to two large clusters, while the four smaller clusters include a mix of advertisers from different categories. The Sankey diagram in Figure \ref{fig:sankei_distance} shows the cluster assignments. Categories that are close in the visualization share similar cluster assignment patterns. Interestingly, we find, that using distance-based clustering, connected advertisers show similar CPC development despite belonging to different categories. For example, ``Sport'' and ``Games'' or ``E-commerce \& Shopping'' and ``Lifestyle'' are mainly located in cluster 1. This indicates dependencies in the bidding mechanism that go beyond classical product categories and expected patterns. We find that products may be more correlated to products from other categories in terms of their CPC development which is an important finding for choosing the right multivariate forecasting approach.

For k-Means clustering based on extracted features, the elbow method suggests seven clusters. In contrast to distance-based clustering, the advertiser categories are not strongly represented. The resulting clusters consist of mixtures of smaller advertiser subgroups, with most categories distributed across multiple clusters. A possible explanation is that the high-level descriptive features used here do not the capture local patterns of the time series that are important for comparing CPC development. The clustering assignment is less interpretable than for distance-based clustering. It also performs worse in terms of multivariate prediction results, indicating that the distance-based clustering is the more meaningful assignment.




\subsection{Model Explainability}\label{section4.3}

We analyze the feature importance of the TFT model by interpreting its Variable Selection Networks (VSNs), which produce sparse weights for static, encoder, and decoder inputs. Additionally, all attention heads of the TFT share the same weights, allowing overall attention to be computed by aggregating individual heads.

\begin{figure*}[!t]
\centering
\includegraphics[width=14.2cm]{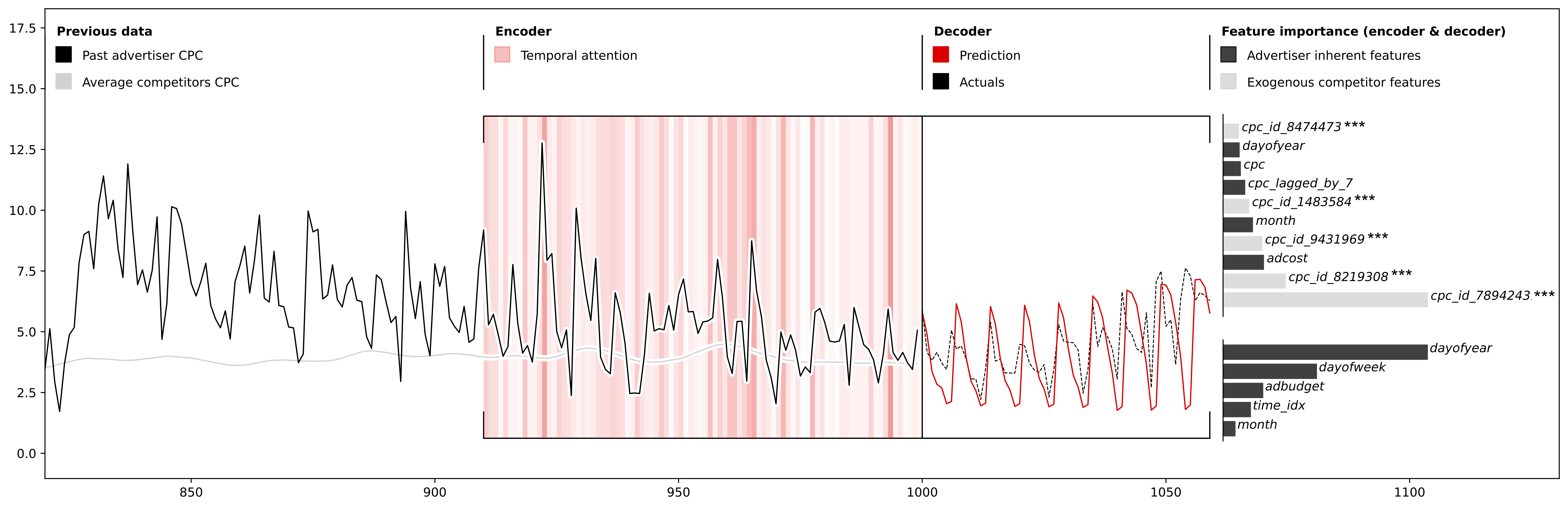}
\includegraphics[width=14.2cm]{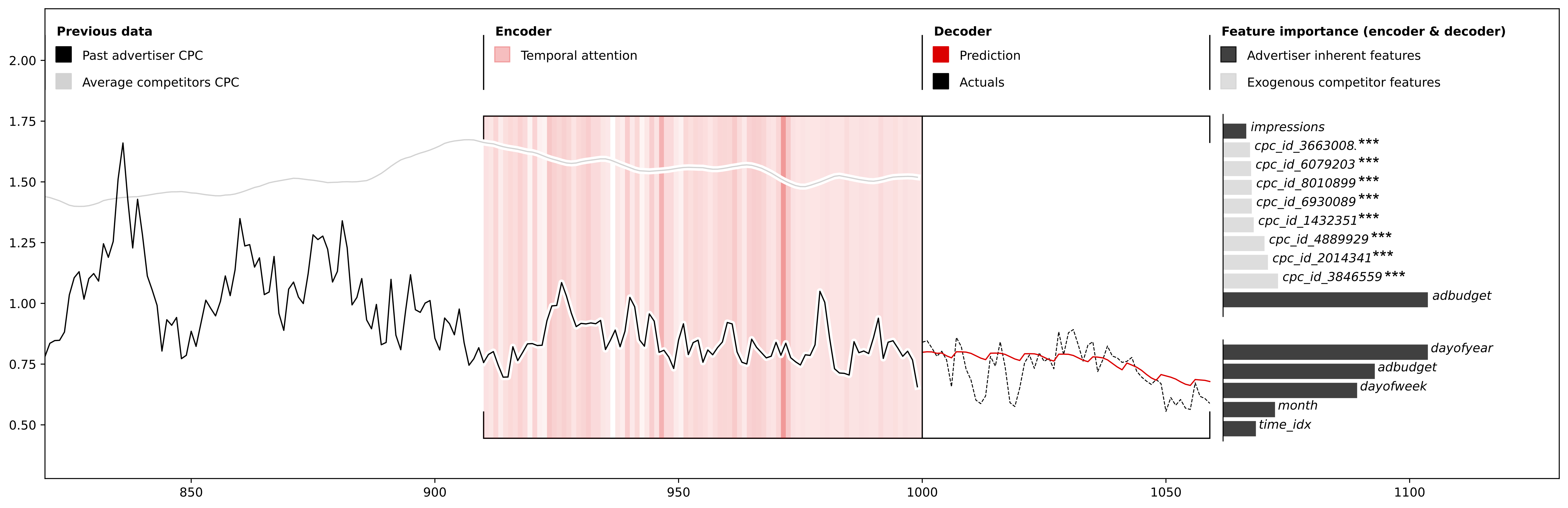}
    \caption{Exemplary 60-day predictions of our proposed model for two selected advertisers showing different underlying CPC structures. Temporal attention to the encoder time steps is highlighted in red, with encoder (top) and decoder (bottom) feature importance shown as bar charts. The average CPC of advertisers in the same clusters is plotted as a grey line.}
 \label{fig:pred_example}
\end{figure*}

Figure \ref{fig:pred_example} demonstrates how the TFT predicts a 60-day CPC forecast for two example advertisers. Temporal attention is visualized as a heatmap behind the time stamps, while encoder and decoder feature importance is shown in sorted bar charts. The average CPC of advertisers within the same cluster is also plotted. The first example shows strong weekly seasonality in CPC. The encoder's importance highlights one competing advertiser as highly influential, while the advertiser's own features, such as CPC and 7-day lagged CPC, are relatively unimportant. This suggests the model is learning level and seasonality from competing advertisers' data. The prediction accurately captures both short-term fluctuations and long-term trends. The second example lacks strong weekly seasonality. Here, the encoder and decoder importances indicate that the monthly advertising budget is a key factor. The TFT prediction reflects a significant future decrease in CPC, likely driven by changes in the planned advertising budget, and the model accurately translates this information into a smooth, long-term decline in CPC.

\begin{figure*}[!t]
\centering
\includegraphics[width=14.2cm]{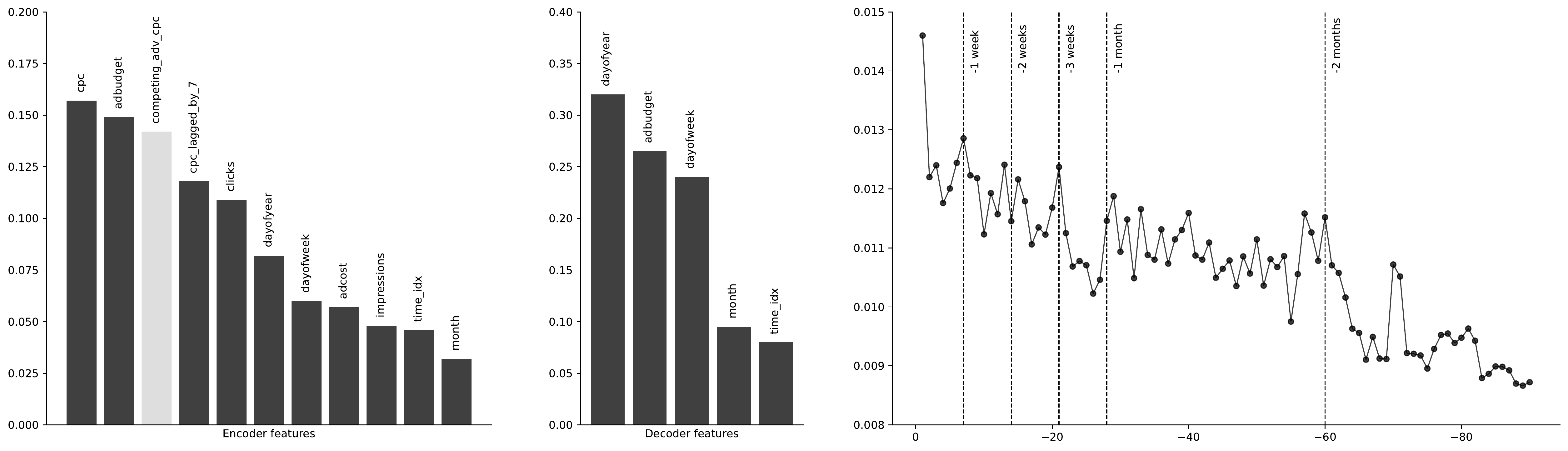}
    \caption{Average feature importance and temporal attention across all advertiser models. Weights dedicated to competing advertisers' CPC features are summed up to one bar in the encoder, highlighted in grey.}
 \label{fig:mean_feat_imp}
\end{figure*}

Figure \ref{fig:mean_feat_imp} shows the average feature importance and temporal attention across all advertiser models. The encoder importance assigned to competing advertisers' CPCs is summed up in one bar. Unsurprisingly, the past values of the target variable, CPC, are most important for the encoder. The advertising budget and competing advertisers also dominate the variable selection for the encoder, followed by the day of the week, which supports modeling weekly seasonalities often present in the data. The decoder outputs the importance of features for which future values are known, with most weight assigned to the day of the year, likely due to special days like Christmas when CPCs spike. The advertising budget and day of the week are also critical for determining the prediction's level and seasonality.

Lastly, Figure \ref{fig:mean_feat_imp} includes the average temporal attention to the time steps in the encoder. Apart from the last available time step, which receives the highest importance, the weights decrease relatively consistently as one moves back in time. We observe weight peaks around time steps representing seasonality cycles, such as every seven days over the first month and roughly every new month in the long term.

\section{Policy Experiment}
\subsection{Robustness to Industry Policy Changes}

A common challenge in deep learning with long time-series data is the presence of large distortions caused by one-time external events, such as economic crises \cite{chatzis2018forecasting}. Significant changes in the data can hinder deep learning models from detecting consistent patterns, leading to lower predictive performance. The timeframe for our study includes the onset of the COVID-19 pandemic, during which Germany, the focus of our geographical data, implemented its first public restrictions, including contact limitations and the cancellation of major events, beginning in March 2020. The travel and tourism sector was particularly impacted by this crisis, with customers delaying or canceling their summer vacation plans amid growing uncertainty about further restrictions \cite{vskare2021impact}.

We assessed the robustness of our proposed model, which leverages information from related advertisers both within and outside their industry. We selected all advertisers from the ``Travel \& Tourism" category within our dataset and tested the performance of two models: the best-performing model from our previous tests that incorporates competing advertisers (TFT.multivar.comp.dist), and a simpler version that only considers each advertiser's characteristics, such as past CPC, clicks, and budget (TFT.multivar). We examined three forecasting horizons: a period before the pandemic's onset in Germany (September and October 2019) to evaluate model performance when trained on stable data; a period immediately after the pandemic's onset (May and June 2020) to assess the impact of the crisis on the online advertising market as reflected in both training data and the 90-day encoder window; and a period a few months after the pandemic outbreak (September and October 2020) to compare how well the models adapt to the new situation in terms of predictive performance.

\begin{figure*}[!t]
\centering
\includegraphics[width=14.2cm]{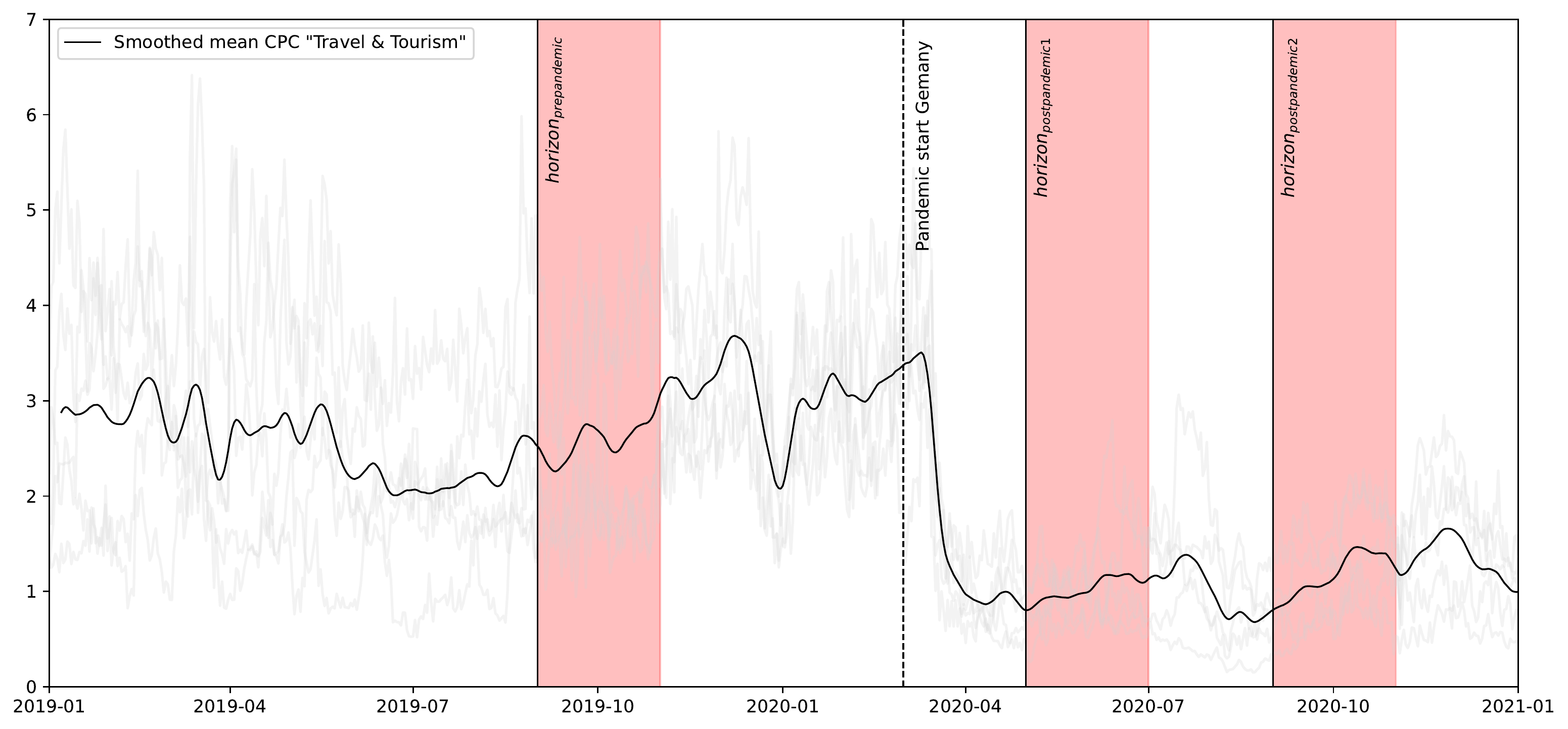}
    \caption{Pandemic impact on the average CPC of six advertisers from ``Travel \& Tourism" with highlighted forecasting horizons to test model performance at different stages of the crisis.}
 \label{fig:pred-example-univariate}
\end{figure*}

Figure \ref{fig:pred-example-univariate} displays the three forecast horizons and the average CPC development for selected advertisers in the ``Travel \& Tourism" category. Notably, we observed a sudden drop in CPC immediately following the start of the pandemic, indicating that advertising became significantly cheaper. CPCs remained low throughout the first pandemic year in Germany. Further analysis revealed that advertisers responded to the pandemic by substantially reducing their advertising budgets. The lowered advertising costs may reflect Google's response. Shortly afterward, advertisers increased their budgets again in the summer of 2020 to nearly pre-pandemic levels, but the conversion to clicks remained high, resulting in consistently lower CPCs throughout the first pandemic year.

\begin{table*}[!t]
\fontsize{7.5}{9}\selectfont
\centering
    {\renewcommand{\arraystretch}{1.5} 
        \begin{tabular}{w{l}{4.5cm}w{c}{2.5cm}w{c}{2.5cm}}
        \hline
          \textbf{Horizon} & \textbf{TFT.multivar} & \textbf{TFT.multivar.comp.dist} \\
          \hline
            Pre-Pandemic \newline (2019-09 and 2019-10) & $0.244 \pm 0.193$ & $0.227 \pm 0.124$ \\
            Post-Pandemic 1 \newline (2020-05 and 2020-06) & $0.750 \pm 0.544$ & $0.687 \pm 0.413$\\
            Post-Pandemic 2 \newline (2020-09 and 2020-10) & $0.528 \pm 0.339$ & $0.329 \pm 0.162$\\
            \hline
        \end{tabular}
        
    }
\vspace{0.2cm}
\caption{Performance improvement measured by mean
SMAPE across six advertisers from ``Travel \& Tourism'' over three forecasting horizons by including clustered competitors' data as covariates}
\label{policy_results}
\end{table*}

Table \ref{policy_results} presents the results of our robustness experiment. In the first forecasting horizon, the difference in performance between the models aligns with our overall results (as shown in Table \ref{table_results}): The TFT model trained with competing advertisers (TFT.multivar.comp.dist) achieves a slightly better SMAPE but remains at a similar performance level as the multivariate TFT model that only uses the advertiser's inherent features (TFT.multivar). In the second horizon, both models exhibit significantly worse performance. The change in CPC levels within the encoder window distorts the predictions, although the TFT.multivar.comp.dist still achieves a better score. In the final horizon, our proposed model demonstrates better adaptability and performance when trained on long data series that include significant changes in the target variable's past development. Incorporating competing advertisers as features in the model training helps capture the interconnected patterns of budget, ad cost, clicks, and resulting CPC, making it less vulnerable to large changes in the training data.

\subsection{Comparison with Google Keyword Planning Tool}
Google Keyword Planning is an online tool designed to predict future CPC for a given keyword based on an advertiser's budget input. However, the models presented and tested in this paper aim to predict an advertiser's total CPC across all keywords bid on. Compared to Google’s tool, our approach supports strategic online marketing decisions regarding budget allocation—specifically, how to maximize clicks while minimizing relative costs (CPC). Our model can answer this question by simulating different future scenarios based on user-defined budget allocation strategies.

Our approach offers two key technical advantages over the Google tool: First, our predictions are made on a daily basis, while the Keyword Planner provides weekly predictions. This allows us to capture greater detail and more accurately represent the strong weekly seasonality present in most CPC time series. Second, our application of the TFT model is interpretable, providing detailed statistics on feature importance and temporal attention alongside the raw forecasts. This enables users to understand seasonality, sensitivity to special days like Christmas, and most importantly, the sensitivity to monthly budget changes. Our model not only delivers accurate forecasts but also provides advertisers with a tool to better understand the drivers of their individual market and identify their direct competitors.

\section{Conclusion}
\subsection{Summary}
This study aimed to forecast online advertising cost-per-click (CPC) across various industries and time horizons using a large-scale paid search dataset. We employed a range of time-series forecasting techniques, including statistical methods, machine learning models, and deep learning architectures. Our experimentation included adding features from the advertiser and similar advertisers to establish multivariate baseline configurations. Three time-series clustering approaches were utilized to identify relevant covariates within the expansive advertiser landscape.

Our findings indicate that the Temporal Fusion Transformer (TFT) architecture, using CPC data from related advertisers within clusters identified through distance-based clustering, achieved the best performance across all model and clustering configurations over three forecasting horizons. This supports the notion that deep learning models outperform statistical methods when applied to large-scale multivariate time-series data \cite{cerqueira2022case}. Additionally, we interpreted the model predictions by analyzing feature importance and temporal attention, demonstrating how the model captures weekly and monthly seasonality as well as CPC information from related advertisers within the same cluster, reflecting the non-linear impact of budget levels on CPC.

Moreover, our approach proved robust even in the face of significant changes in the online advertising market, such as those triggered by the COVID-19 pandemic. Compared to model configurations using the same architecture but limited to the focal advertiser's CPC information, our method consistently delivered superior performance across various time horizons.

This study contributes to the online advertising community by proposing a scalable technique for selecting relevant covariates from a wide pool of advertisers, thereby improving multi-horizon prediction performance. Our approach allows individual advertisers to input future budget plans and receive accurate long-term forecasts based on patterns observed in the past. The components of our proposed forecasting pipeline can also be applied to other tasks, such as categorizing advertisers not natively assigned to a category or accurately filling gaps in advertisers' time-series data.



\subsection{Future Works}

The strong correlations of CPC changes across industries suggest both opportunities for deeper research and a need for practitioners to broaden their perspectives. Advertisers should not only monitor CPC developments and predictions within their sector but also explore non-obvious connections between advertisers from different industries, as identified through our distance-based clustering approach. Future research at the keyword level is necessary to further investigate advertising connections across industries.


Recent advances in Graph Neural Networks (GNNs), which use either static or dynamic temporal graph representations of a network of series \cite{wu2020, cui2021}, appear well-suited for modeling the competitive landscape of online advertising. It would be valuable to compare performance using graph structures based on distances in multivariate series, as in this study, but with finer granularity. Additionally, constructing graph structures based on keyword data, where each node represents an advertiser and edge weights represent shared keywords, could provide further insights. Future research could also explore the incorporation of recent advances in large language models (LLMs) or foundation models for time-series forecasting \cite{liang2024foundation, ye2024survey}. These models, with their extensive pre-trained knowledge and ability to fine-tune for specific tasks, could potentially enhance the predictive performance and interpretability of time-series forecasts, further advancing the capabilities of online advertising analytics.




%
\printbibliography

@article{hochreiter1997long,
  title     = {{Long short-term memory}},
  author    = {Hochreiter, Sepp and Schmidhuber, J{\"u}rgen},
  journal   = {Neural computation},
  volume    = {9},
  number    = {8},
  pages     = {1735--1780},
  year      = {1997},
  publisher = {MIT Press}
}

@inproceedings{yuan2013,
  author  = {Shuai Yuan and Jun Wang and Xiaoxue Zhao},
  year    = {2013},
  month   = {August},
  pages   = {1-8},
  title   = {{Real-time Bidding for Online Advertising: Measurement and Analysis}},
  booktitle = {Seventh International Workshop on Data Mining for Online Advertising},
  doi     = {10.1145/2501040.2501980}
}

@article{bandara2020,
  title={{Forecasting across time series databases using recurrent neural networks on groups of similar series: A clustering approach}},
  author={Kasun Bandara and Christoph Bergmeir and Slawek Smyl},
  journal={{Expert Systems with Applications}},
  volume={140},
  number={112896},
  year={2020}
}

@inproceedings{chen2016,
  author  = {Junxuan Chen and Baigui Sun and Hao Li and Hongtao Lu and Xian-Sheng Hua},
  year    = {2016},
  month   = {October},
  pages   = {811-820},
  title   = {{Deep CTR Prediction in Display Advertising}},
  booktitle = {MM '16: Proceedings of the 24th ACM international conference on Multimedia}
}

@article{Spiliotis2021,
  title   = {{Comparison of statistical and machine learning methods for daily SKU demand forecasting}},
  author  = {Evangelos Spiliotis and Spyros Makridakis and Artemios-Anargyros Semenoglou and Vassilios Assimakopoulos },
  journal = {{Operation Research}},
  volume  = {22},
  pages = {3037-3061},
  year    = {2021}
}

@inproceedings{Berndt1994,
author = {Donald J. Berndt and James Clifford},
title = {Using Dynamic Time Warping to Find Patterns in Time Series},
year = {1994},
booktitle = {KDD '94, 10th ACM SIGKDD International Conference on Knowledge Discovery \& Data Mining},
pages = {359–370}
}

@misc{WordStream2022,
  title     = {{Google Ads Benchmarks for YOUR Industry [Updated!]}},
  url       = {https://www.wordstream.com/blog/ws/2016/02/29/google-adwords-industry-benchmarks},
  journal   = {WordStream},
  author = {{Mark Irvine}},
  year = {2022}
}

@article{cerqueira2022case,
  title={A case study comparing machine learning with statistical methods for time series forecasting: size matters},
  author={Cerqueira, Vitor and Torgo, Luis and Soares, Carlos},
  journal={Journal of Intelligent Information Systems},
  volume={59},
  number={2},
  pages={415--433},
  year={2022},
  publisher={Springer}
}

@article{cui2021,
  title={METRO: a generic graph neural network framework for multivariate time series forecasting},
  author={Cui, Yue and Zheng, Kai and Cui, Dingshan and Xie, Jiandong and Deng, Liwei and Huang, Feiteng and Zhou, Xiaofang},
  journal={Proceedings of the VLDB Endowment},
  volume={15},
  number={2},
  pages={224--236},
  year={2021},
  publisher={VLDB Endowment}
}

@misc{dentsu2023,
  title = {{Global Ad Spend Forecast December 2023}},
  url = {https://insight.dentsu.com/ad-spend-dec-2023/},
  author = {{Dentsu}},
  year = {2023}
}

@misc{googleadsense2022,
  title     = {{How AdSense works}},
  url       = {https://support.google.com/adsense/answer/6242051?hl=en},
  journal   = {Google},
  publisher = {Google},
  author    = {Google},
  year      = {2022}
}

@article{huang2016,
  title = {Time series k-means: A new k-means type smooth subspace clustering for time series data},
  journal = {Information Sciences},
  volume = {367-368},
  pages = {1-13},
  year = {2016},
  issn = {0020-0255},
  author = {Xiaohui Huang and Yunming Ye and Liyan Xiong and Raymond Y.K. Lau and Nan Jiang and Shaokai Wang},
}

@article{lim2021survey,
	year = 2021,
	month = {feb},
	publisher = {The Royal Society},
	volume = {379},
	number = {2194},
	pages = {20200209},
	author = {Bryan Lim and Stefan Zohren},
	title = {Time-series forecasting with deep learning: a survey},
	journal = {Philosophical Transactions of the Royal Society A: Mathematical, Physical and Engineering Sciences}
}

@article{lim2021temporal,
  title = {Temporal Fusion Transformers for interpretable multi-horizon time series forecasting},
  journal = {International Journal of Forecasting},
  volume = {37},
  number = {4},
  pages = {1748-1764},
  year = {2021},
  issn = {0169-2070},
  author = {Bryan Lim and Sercan. Arık and Nicolas Loeff and Tomas Pfister}
}

@article{makridakis2018,
  title   = {{Statistical and Machine Learning forecasting methods: Concerns and ways forward}},
  author  = {Makridakis, Spyros and Spiliotis, Evangelos and Assimakopoulos, Vassilios},
  journal = {{PLoS ONE}},
  volume  = {13},
  number  = {3},
  pages   = {e0194889},
  year    = {2018}
}

@article{petitjean2011,
  title = {A global averaging method for dynamic time warping, with applications to clustering},
  journal = {Pattern Recognition},
  volume = {44},
  number = {3},
  pages = {678-693},
  year = {2011},
  issn = {0031-3203},
  author = {François Petitjean and Alain Ketterlin and Pierre Gançarski},
}

@inproceedings{wu2020,
  title={Connecting the dots: Multivariate time series forecasting with graph neural networks},
  author={Wu, Zonghan and Pan, Shirui and Long, Guodong and Jiang, Jing and Chang, Xiaojun and Zhang, Chengqi},
  booktitle={Proceedings of the 26th ACM SIGKDD International Conference on Knowledge Discovery \& Data Mining},
  pages={753--763},
  year={2020}
}

@article{yang2022,
  title = {Time-varying effects of search engine advertising on sales–An empirical investigation in E-commerce},
  journal = {Decision Support Systems},
  volume = {163},
  pages = {113843},
  year = {2022},
  issn = {0167-9236},
  author = {Yanwu Yang and Kang Zhao and Daniel Dajun Zeng and Bernard Jim Jansen},
}

@article{zia2019search,
  title={Search advertising: Budget allocation across search engines},
  author={Zia, Mohammad and Rao, Ram C},
  journal={Marketing Science},
  volume={38},
  number={6},
  pages={1023--1037},
  year={2019},
  publisher={INFORMS}
}

@techreport{greenwood2021you,
  title={You will: A macroeconomic analysis of digital advertising},
  author={Greenwood, Jeremy and Ma, Yueyuan and Yorukoglu, Mehmet},
  year={2021},
  institution={National Bureau of Economic Research}
}

@article{evans2008economics,
  title={The economics of the online advertising industry},
  author={Evans, David S},
  journal={Review of Network Economics},
  volume={7},
  number={3},
  year={2008},
  publisher={De Gruyter}
}

@article{shen2023price,
  title={Price and advertising competition in an online marketplace: The tradeoff between quality and cost},
  author={Shen, Yuelin},
  journal={Electronic Commerce Research and Applications},
  volume={60},
  pages={101276},
  year={2023},
  publisher={Elsevier}
}

@article{rutz2011modeling,
  title={Modeling indirect effects of paid search advertising: Which keywords lead to more future visits?},
  author={Rutz, Oliver J and Trusov, Michael and Bucklin, Randolph E},
  journal={Marketing Science},
  volume={30},
  number={4},
  pages={646--665},
  year={2011},
  publisher={INFORMS}
}

@article{bagwell2007economic,
  title={The economic analysis of advertising},
  author={Bagwell, Kyle},
  journal={Handbook of Industrial Organization},
  volume={3},
  pages={1701--1844},
  year={2007},
  publisher={Elsevier}
}

@article{r2020advertising,
  title={Advertising and COVID-19},
  author={Charles Taylor},
  journal={International Journal of Advertising},
  volume={39},
  number={5},
  pages={587--589},
  year={2020},
  publisher={Taylor \& Francis}
}

@article{chatzis2018forecasting,
  title={Forecasting stock market crisis events using deep and statistical machine learning techniques},
  author={Chatzis, Sotirios P and Siakoulis, Vassilis and Petropoulos, Anastasios and Stavroulakis, Evangelos and Vlachogiannakis, Nikos},
  journal={Expert systems with applications},
  volume={112},
  pages={353--371},
  year={2018},
  publisher={Elsevier}
}

@article{liang2024foundation,
  title={Foundation models for time series analysis: A tutorial and survey},
  author={Liang, Yuxuan and Wen, Haomin and Nie, Yuqi and Jiang, Yushan and Jin, Ming and Song, Dongjin and Pan, Shirui and Wen, Qingsong},
  journal={arXiv preprint arXiv:2403.14735},
  year={2024}
}

@article{ye2024survey,
  title={A Survey of Time Series Foundation Models: Generalizing Time Series Representation with Large Language Mode},
  author={Ye, Jiexia and Zhang, Weiqi and Yi, Ke and Yu, Yongzi and Li, Ziyue and Li, Jia and Tsung, Fugee},
  journal={arXiv preprint arXiv:2405.02358},
  year={2024}
}

@inproceedings{wurfel2021online,
  title={Online advertising revenue forecasting: An interpretable deep learning approach},
  author={Würfel, Max and Han, Qiwei and Kaiser, Maximilian},
  booktitle={2021 IEEE International Conference on Big Data (Big Data)},
  pages={1980--1989},
  year={2021},
  organization={IEEE}
}

@article{vskare2021impact,
  title={Impact of COVID-19 on the travel and tourism industry},
  author={{\v{S}}kare, Marinko and Soriano, Domingo Riberio and Porada-Rocho{\'n}, Ma{\l}gorzata},
  journal={Technological Forecasting and Social Change},
  volume={163},
  pages={120469},
  year={2021},
  publisher={Elsevier}
}

\end{document}